\documentclass[conference]{IEEEtran}

\usepackage{times}
\usepackage{graphicx}
\usepackage{wrapfig}
\usepackage{amsmath}
\usepackage{amssymb}
\usepackage{multicol,lipsum}

\usepackage[numbers]{natbib}
\usepackage{multicol}
\usepackage{booktabs} 
\usepackage{array} 
\usepackage[colorlinks=true]{hyperref}
\usepackage{balance}

\usepackage{tabularx} 
\usepackage{ifthen}
\usepackage{xcolor}

\usepackage{enumitem}
\usepackage{xspace}
\usepackage{float}
\usepackage{dblfloatfix} 
\usepackage{algorithm}
\usepackage[noend]{algorithmic}
{}
{}
{}
\IEEEoverridecommandlockouts
\usepackage{cite}
\usepackage{amsmath,amssymb,amsfonts}
\usepackage{algorithmic}
\usepackage{graphicx}
\usepackage{textcomp}
\usepackage{xcolor}
\def\BibTeX{{\rm B\kern-.05em{\sc i\kern-.025em b}\kern-.08em
    T\kern-.1667em\lower.7ex\hbox{E}\kern-.125emX}}

\usepackage{cuted}
\title{Robot Learning with Super-Linear Scaling}

\author{
\textbf{Marcel Torne}$^{1,3,*}$ \quad \textbf{Arhan Jain}$^{2,*}$ \quad \textbf{Jiayi Yuan}$^{2,*}$ \quad \textbf{Vidaaranya Macha}$^{2,*}$ 
\quad \textbf{Lars Ankile}$^{1}$ \\ \textbf{Anthony Simeonov}$^{1}$ \textbf{Pulkit Agrawal}$^{1}$ \quad \textbf{Abhishek Gupta}$^{2}$ \\
$^1$Massachusets Institute of Technology \quad $^2$University of Washington \quad $^3$ Stanford University 
\\
}

\newcommand{\MethodName}{\text{\textbf{CASHER}}\xspace}

\begin{document}
\maketitle

\let\thefootnote\relax\footnotetext{* Equal contribution}


\begin{abstract}
    Scaling robot learning requires data collection pipelines that scale favorably with human effort. In this work, we propose \textit{Crowdsourcing and Amortizing Human Effort for Real-to-Sim-to-Real}(\MethodName), a pipeline for scaling up data collection and learning in simulation where the performance scales superlinearly with human effort. The key idea is to crowdsource digital twins of real-world scenes using 3D reconstruction and collect large-scale data in simulation, rather than the real-world. Data collection in simulation is initially driven by RL, bootstrapped with human demonstrations. As the training of a generalist policy progresses across environments, its generalization capabilities can be used to replace human effort with model-generated demonstrations. This results in a pipeline where behavioral data is collected in simulation with continually reducing human effort. We show that \MethodName demonstrates zero-shot and few-shot scaling laws on three real-world tasks across diverse scenarios. We show that \MethodName enables fine-tuning of pre-trained policies to a target scenario using a video scan without any additional human effort.
    Project website: \url{https://casher-robot-learning.github.io/CASHER/}
\end{abstract}


\section{Introduction}
\label{sec:introduction}


Robot learning has the potential to revolutionize decision-making for robots by leveraging data to learn behaviors deployable in unstructured environments, showing generalization and robustness. Critical to the success of robot learning, beyond the algorithms and model architectures, is the training data. As in most machine learning, getting the ``right" type, quality, and quantity of data holds the key to generalization. Robot learning is still grappling with the question of what the right type of data and how to obtain it at scale. The type of data we can train on is inherently tied to the abundance of this data - good data is both high-quality and abundant. This paper proposes a system for obtaining this diverse, high-quality data at superlinear scale with sublinear human effort. 

Unlike vision and language, data for learning is not available passively - there are relatively few robots that are already finding use in the world. This makes applying the same recipes we did in vision and language challenging, necessitating more careful consideration of how and where this data comes from. One option is to rely on teleoperation to collect this data. This approach is inherently limited by human effort, since the cost to collect data scales linearly with human involvement. Recent work \citep{khazatsky2024droid, padalkar2023open, brohan2022rt} has attempted to scale the amount of teleoperation data however the quantity of data collected is still orders of magnitude smaller than the scale at which vision and language models show emergent capabilities. 


So where might we find data that scales superlinearly with human effort? Simulation offers a potential solution, at face value providing free data up to the limit of computing. However, this hides a significant cost - scene, task, and reward creation per domain is non-trivial, and even with scenes generated, behaviors are costly to obtain. This suggests that despite the promise, simulation data isn't quite free of cost, and requires considerable amounts of human efforts for content and behavior creation per environment. While it is possible to generate random environments procedurally, generating thousands of environments randomly is unlikely to cover the distribution of "natural environments", and generating behaviors randomly is unlikely to lead to success. 

\begin{figure*}[t!]
    \centering
    \includegraphics[width=0.9\linewidth]{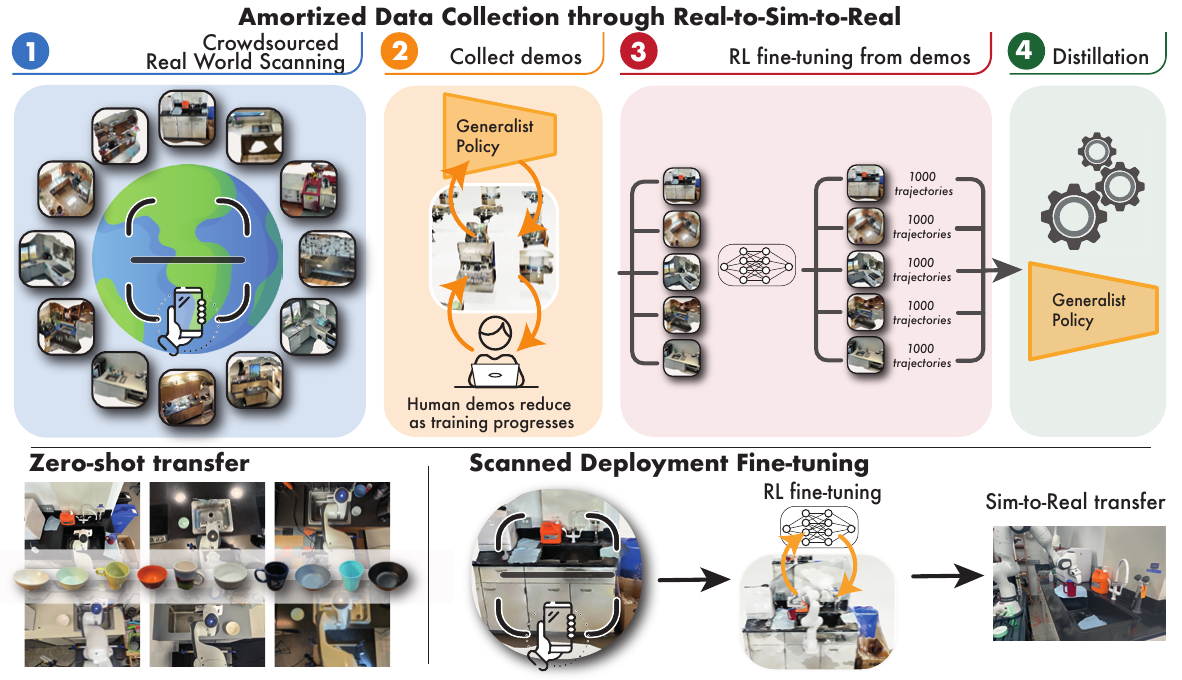}
    \caption{\footnotesize{Overview of \MethodName, we propose a system for training generalist policies leveraging real-to-sim simulation on 
    crowdsourced scans. These have zero-shot transfer and scanned fine-tuning capabilities.} }
    \label{fig:mainfigure}
\end{figure*}

In this work, we propose a method to scale up continual data collection, ensuring human effort amortizes sublinearly with the number of environments. Our key idea is to leverage simulation for data scaling without the corresponding increase in content and behavior creation effort. For content scaling, we utilize 3D reconstruction methods, shifting the burden from designers to non-expert users and cheap data collection. For behavior generation, we employ techniques that leverage model generalization to reduce the required human data over time. The insight is that as we go across many simulated environments, models will show some levels of generalization. This generalization can be leveraged to continually reduce the amount of human data needed as new environments are encountered. \MethodName creates a data flywheel, where data begets more data through model generalization. 


Our contributions include 1) a novel continual data collection system based on real-to-sim-to-real for training generalist policies, 2) a novel scanned deployment fine-tuning technique for improving the accuracy of a generalist policy on a target environment without additional human demonstrations, 3) a detailed analysis of the scaling laws for zero-shot performance of our generalist policies, 4) evaluation of the few-shot performance of the resulting generalist policies.

\section{Related Work}
\label{sec:related_work}

\textbf{Large Scale Data Collection for Robotics}: Learning from real-world demonstrations has proven effective \citep{chi2023diffusion,zhao2023learning, octo_2023}. To facilitate this, various studies have focused on improving hardware to ease the data collection process for teleoperators \citep{zhao2023learning, chi2024universal, wu2023gello}. Efforts have also scaled up the volume of data from real-world demonstrations \citep{padalkar2023open, khazatsky2024droid, brohan2022rt}, staying nevertheless in the low-data regime. Moreover, real-world data collection is costly, requiring expert supervision and physical robots, which limits scalability. \MethodName, instead, trains entirely in simulation, using real-world scans obtained via standard smartphones. Additionally, while traditional teleoperation data collection scales linearly with human effort, \MethodName reduces the human effort needed for subsequent learning steps by leveraging the knowledge acquired during training.

\textbf{Autonomous Learning}: To improve scalability of robot learning  and reduce the amount of human demonstrations required, the field has explored autonomous data collection and learning methods. One approach is reinforcement learning (RL) in the real world \citep{luo2024serl, levine2018learning}, but the standard RL techniques' need for resets poses scalability issues, as it requires either human supervision or substantial engineering efforts for automating resets. Reset-free reinforcement learning \citep{balsells2023autonomous, yang2023robot, sharma2023self, gupta2021reset} offers a promising alternative, but it still requires occasional human intervention and struggles with high sample complexity for learning more challenging tasks, making it hard to learn in the real-world.
Autonomous learning in the real world presents significant challenges that are mitigated in simulation, where resets are manageable and data collection is more abundant. In \MethodName, we exploit these advantages of simulation while minimizing the sim-to-real gap through real-to-sim scene transfers.
Continual learning also faces challenges, such as catastrophic forgetting, as discussed in prior work \citep{lesort2020continual}. We address this by decoupling the policy used to generate trajectories, which is fine-tuned with RL, from the final generalist policy, which is trained with imitation learning over the entire dataset.

\textbf{Procedural and Synthetic Data Generation}: Creating realistic environments for robot learning in simulation is a significant challenge. To address this, prior work has proposed using large language models (LLMs) or heuristics to generate scene plans resembling the real world \citep{wang2023gensim, deitke2022, robocasa2024}, or utilizing real-world scans to replicate actual scenes \citep{deitke2023phone2proc, chen2024urdformer}. Despite reducing human involvement, these methods often produce scenes that are unrealistic in appearance or object distribution, such as failing to accurately simulate real-world clutter.
Generating procedurally accurate training environments remains an open challenge. However, extracting digital twins from the real world mitigates this issue, as scans reflect the actual test distribution. Relevant to our work, \citep{xia2024video2game} automates the creation of simulatable environments from real-world scans, which could be integrated into our pipeline to scale up environment crowdsourcing.
Once the environments are available, generating valid robot trajectories that solve the task is another challenge. An option becomes procedurally generating the motions using motion planning techniques \citep{ha2023scaling}. However, these techniques require some assumptions beforehand.



\textbf{Real-to-Sim-to-Real Transfer for Robotics}: Real-to-sim-to-real techniques have proven effective in learning robust policies for specific scenarios with minimal human supervision \citep{torne2024reconciling, wang2023real2sim2real}. However, these policies often fail to generalize to different scenarios, requiring significant human effort for each new environment. In this work, we address this limitation by learning generalist policies through a novel technique that amortizes the number of human demonstrations through training.
Other research has tackled various challenges in real-to-sim-to-real, such as enhancing simulator accuracy with real-world interaction data \citep{memmel2024asid, ramos2019bayessim, chebotar2019closing}, and automatically generating articulations from images \citep{chen2024urdformer, jiang2022ditto, nie2022sfa}. These complementary advancements make simulators more realistic and could reduce human effort further in \MethodName.
Additionally, real-to-sim techniques have shown promise in their use for simulated evaluation of real-world policies \citep{li24simpler}.

\section{Amortized Data Scaling for Learning Generalist Policies through Real-to-Sim-to-Real}
\label{sec:method}

This work presents \MethodName, a pipeline for large-scale continual data collection for robotic manipulation. The primary challenge for data scaling in the realm of robotics is the absence of ``passive", easy-to-collect data from naturally occurring, inadvertent sources, as is common in vision and language. While procedural generation in simulation can provide large amounts of data, the distribution and diversity of the data does not overlap with real-world environments. In this work, we argue that a multi-task, multi-environment real-to-sim-to-real pipeline can enable large-scale data generation, by leveraging model generalization to scale human-effort sublinearly as increasing numbers of environments are encountered. This is opposed to typical human teleoperated data collection that requires considerable expertise, physical infrastructure and suffers from linear scaling in human effort. This  approach enables the scaling laws necessary for large scale data collection and training of robotic foundation models, showing non-trivial zero-shot generalization performance as well as cheap and efficient fine-tuning in new environments. \MethodName consists of three elements - 1) fast, accessible digital twin generation with 3-D reconstruction methods, 2) multi-environment model learning that amortizes the data collection process through autonomous data collection and model generalization, 3) efficient fine-tuning in new environments using 3-D scans, and minimal human demonstrations. 


\subsection{Real-to-Sim Scene Synthesis} 
\label{sec:r2simscene}

Our proposed data collection pipeline adopts a real-to-sim-to-real approach, building digital twins of real-world scenes in simulation and collecting behavioral data in these simulations instead of the real world. This method offers several advantages - 
1) data collection does not require a physical robot setup, and hence can occur in a broader variety of realistic environments 
2) it allows for safe, decentralized, and asynchronous data collection
3) digital twins capture the complexities of real-world scenarios more accurately than procedurally generated simulations. These advantages are crucial to the democratization and scalability of data collection as it is scaled up to thousands of non-experts and real environments beyond the lab. 
We leverage easily accessible mobile software\citep{arcode2022,polycam2020} for scene reconstruction from sequences of images to easily crowdsource simulated environments \footnote{We provide further details about the real-to-sim pipeline in Appendix \ref{apdx:real2sim}, including how to stage these scenes, articulate them quickly and so on.}. These environments indicate the geometry, visuals and physics of diverse real-world scenes in simulation but do not have any demonstrations of the desired optimal behavior. We discuss how this can be obtained efficiently in the following section. 




\subsection{Amortized Data Collection}
\label{sec:autonomousdatacollection}

\begin{figure*}[h!]
    \centering
    
    \includegraphics[width=0.9\linewidth]{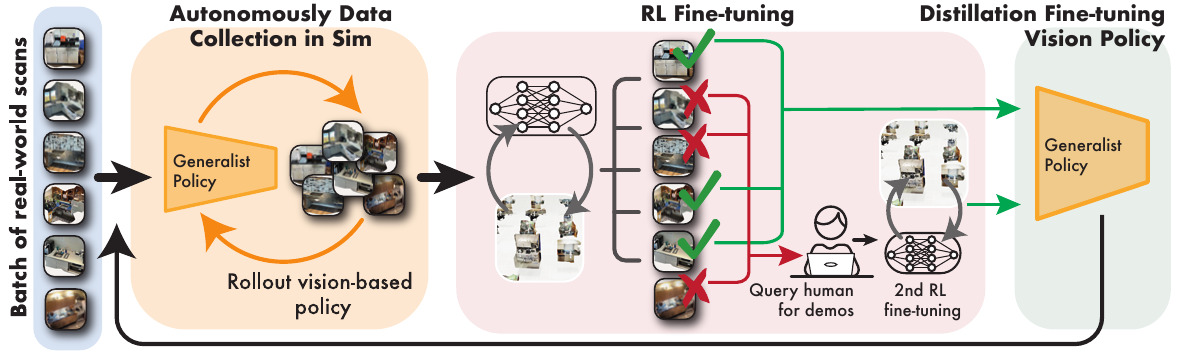}
    \caption{\footnotesize{Overview of the proposed continual data collection system for amortizing human data collection.} }
    \label{fig:continuallearningfigure}
    \vspace{-10pt}
\end{figure*}

Given the diversity of realistic simulation scenes available through the digital twin pipeline outlined in Section~\ref{sec:r2simscene}, learning generalizable decision-making policies requires a large training set of visuomotor trajectories demonstrating optimal behavior for each distinct environment. Two natural alternatives for obtaining these trajectories are: 1) human-provided demonstrations and 2) optimal policies trained via reinforcement learning \footnote{Other techniques such as trajectory optimization or motion planning may be applicable as well}. While tabula-rasa reinforcement learning can provide a robust set of trajectories with extensive state coverage without expensive human intervention, it faces considerable challenges related to exploration and reward design. On the other hand, human demonstrations avoid these issues but are expensive to collect at scale.

A natural solution is to use sparse-reward reinforcement learning bootstrapped with human demonstrations~\citep{torne2024reconciling,hu2023imitation, Rajeswaran-RSS-18}\footnote{We refer readers to Appendix \ref{apdx:multitaskrl} for details of demonstration bootstrapped reinforcement learning}.
This approach balances human effort for data collection and reward specification with state-space coverage. However, scaling it up to hundreds or thousands of scenes becomes tedious, as the required human effort increases linearly with the number of environments. In this work, we learn a generalist multi-environment policy to \emph{amortize} the cost of human data collection across environments. We demonstrate that the capacity of such a multi-environment model to display non-trivial generalization allows the cost of continual human data collection to decrease as the number of training environments increases. 

This system, formally stated in the Appendix Algorithm \ref{alg:amortized}, divides the total number of environments into batches of size $K$. For the first batch of $K$ environments $\mathcal{E}_1, \mathcal{E}_2, \dots, \mathcal{E}_K$, we have a multi-environment visuomotor policy $\pi_G$ randomnly initialized with no generalization capabilities. Thereafter, we initialize it with data from the first $K$ environments, using reinforcement learning bootstrapped with human-provided demonstrations. Demonstration bootstrapped RL produces optimal visuomotor trajectories per environment $\mathcal{D}$, that are then distilled into a single perception-based, generalist multi-environment policy $\pi_G$ with visuomotor policy distillation \citep{chen2023visual} (Appendix \ref{apdx:teacherstudentdistill}). 


While human demonstrations are used to bootstrap the data generation and training of the first iteration of the generalist policy $\pi_G$ on the first $K$ environments, our key insight is that if $\pi_G$ shows non-trivial level of generalization on visuomotor deployment in the next $K$ simulation environments - $\mathcal{E}_{K+1}, \dots, \mathcal{E}_{2K}$, then this policy $\pi_G$ can be used to collect simulated demonstrations $\mathcal{T} = \tau_{K+1,1},\tau_{K+1,2}, \dots, \tau_{2K,N}$ in place of a human demonstrator. We do so by deploying the visuomotor policy $\pi_G(a_t|o_t)$ using perceptual observations $o_t$ such as RGB point clouds, but since we are in the simulation we collect $\mathcal{T}$ with paired data of visual observations $o_t$, actions $a_t$ and low-dimensional privileged Lagrangian state $s_t$. These privileged state-based trajectories enable the usage of efficient demonstration-bootstrapped reinforcement learning of a state-based policy $\pi_s$ rather than operating from high-dimensional perceptual observations. See Eq \ref{rl_finetune} and Appendix \ref{apdx:multitaskrl} for the state-based policy update using PPO \citep{schulman2017proximal} with a BC loss, where $\hat{A}_t$ is the estimator of the advantage function at step $t$ \citep{schulman2017proximal}, and $V_{\phi}$ is the learned value function.

\begin{strip}
\begin{equation}
    \label{rl_finetune}
    \begin{split}
            \pi_s \leftarrow
            \max_{\theta,\phi}
            \alpha \sum_{\mathcal{E}_i\in \{ \mathcal{E}_1, \mathcal{E}_{2}, \dots, \mathcal{E}_{K} \} } \sum_{(s_t, a_t, r_t)\in \mathcal{E}_i(\pi_{\theta_{\text{old}}})}\text{min} (\frac{\pi_\theta(a_t|s_t)}{\pi_{\theta_{\text{old}}}(a_t|s_t)}\hat{A}_t, 
            \text{clip}(\frac{\pi_\theta(a_t|s_t)}{\pi_{\theta_{\text{old}}}(a_t|s_t)}, 1-\epsilon, 1+\epsilon)\hat{A}_t)  \\
            + \beta \sum_{\mathcal{E}_i\in \{ \mathcal{E}_1, \mathcal{E}_{2}, \dots, \mathcal{E}_{K} \} }  \sum_{(s_t, V_t^{\text{targ}})\in \mathcal{E}_i(\pi_{\theta_{\text{old}}})} (V_\phi(s_t) - V_t^\text{targ})^2  
            + \gamma \sum_{(s_i, a_i)\in \mathcal{T}} \log \pi_\theta(a_i|s_i) 
    \end{split}
\end{equation}
\end{strip}


$\mathcal{T}$ can be used to obtain a single robust, state-covering optimal multi-environment policy ${\pi_s}_1(a_t|s_t)$ for all $\mathcal{E}_{K+1}, \dots, \mathcal{E}_{2K}$ via demonstration-bootstrapped reinforcement learning. Nevertheless, in some environments, the policy may still perform poorly due to the occasional low-quality demonstrations from $\pi_G$. To address this, we define the set of environments where ${\pi_s}_1$ achieves below $r$ success rate as $\mathcal{F} \subset \{\mathcal{E}_K, \mathcal{E}_{K+1}, \dots, \mathcal{E}_{2K}\}$. For these environments $\mathcal{F}$, we fall back to querying the human demonstrator for high-quality demonstrations and learn a second state-based policy ${\pi_s}_2(a_t|s_t)$ using demonstration-bootstrapped reinforcement learning on $\mathcal{F}$. 


The two learned policies ${\pi_s}_1$ and ${\pi_s}_2$ can then be used for generating data on $\{\mathcal{E}_K, \mathcal{E}_{K+1}, \dots, \mathcal{E}_{2K}\}\backslash \mathcal{F}$ and $\mathcal{F}$ respectively with these new trajectories being added into $\mathcal{D}$. Then,  a visuomotor policy can be trained by fitting $\mathcal{D}$ on the first $2K$ environments with supervised learning  (see Appendix \ref{apdx:teacherstudentdistill} for implementation details). 

\begin{equation}
    \label{eq_distill}
            \pi_G \leftarrow \max_{\theta} \mathbb{E}_{(o_i, a_i) \sim \mathcal{D}}
            \left[ \log {{\pi_G}_\theta}(a_i|o_i) \right]
\end{equation}

Then the process repeats for the next $K$ environments. As the visuomotor generalist policy $\pi_G$ is trained across more environments, it demonstrates increasingly non-trivial generalization, gradually replacing the human demonstrator in more environments. This reduces the amount of human effort required for data collection as training progresses. Importantly, the generalization across environments does not need to achieve perfect success rates but should be sufficient to bootstrap a demonstration-augmented policy learning algorithm (Equation \ref{rl_finetune}). This suggests an interesting scaling law - data collection becomes more human-efficient as training progresses, eventually becoming self-sustaining. For a detailed outline of the practical data collection pipeline, refer to Algorithm \ref{alg:amortized}. 

\begin{figure*}[ht!]
    \centering
    \includegraphics[width=0.9\linewidth]{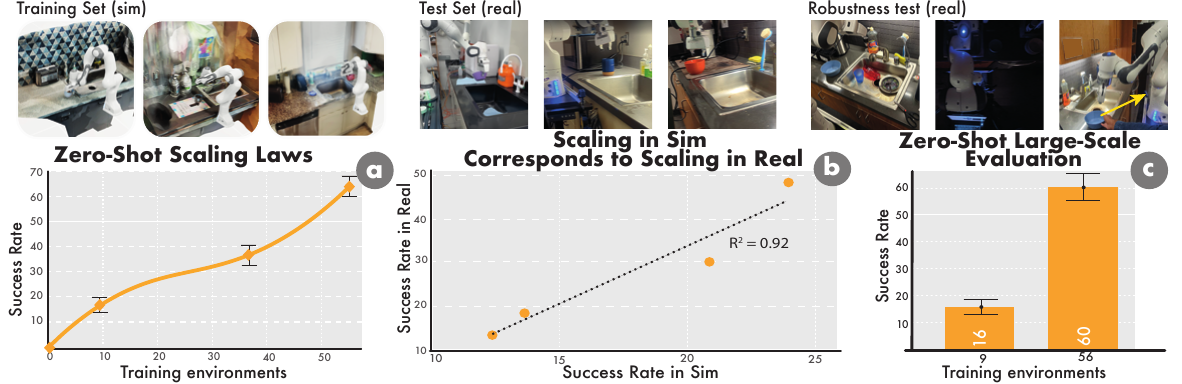}
    \caption{\footnotesize{\emph{a}) \MethodName's zero-shot scaling laws on the task of  \emph{pick and placing bowl/cup/mugs to sinks}; \emph{b}) in the proposed real-to-sim-to-real setup there is a linear relation between performance in sim and performance in real; \emph{c}) evaluation on a broader set of environments confirms the robustness of the zero-shot policies.} }
    \label{fig:zeroshotfig}
\end{figure*}

\subsection{Fine-tuning of Generalist Policies on Deployment}
\label{sec:finetuning}

The generalist policies $\pi_G(a_t|o_t)$ pretrained in Section \ref{sec:autonomousdatacollection}, show non-trivial generalization across environments but may not achieve optimal performance in any one environment upon zero-shot deployment. However, these generalist policies can serve as a starting point for efficient fine-tuning at test time. In this section, we present an alternative for fine-tuning generalist policies $\pi_G(a_t|o_t)$ during deployment. We make the observation that we can follow the same procedure as model-bootstrapped autonomous data collection during training described in Section \ref{sec:autonomousdatacollection}. Given a scanned digital twin $\mathcal{E}_{\text{test}}$ of the testing environment in simulation, the pre-trained multi-environment model $\pi_G(a_t|o_t)$ shows some non-trivial zero-shot generalization, but may not achieve optimal performance in $\mathcal{E}_{\text{test}}$. By executing the visuomotor policy $\pi_G(a_t|o_t)$ in $\mathcal{E}_{\text{test}}$, we collect a dataset of only successful trajectories $\mathcal{T}_{\text{test}}$ consisting of $(o_t, a_t, s_t)$ tuples in simulation, without the need for any external human intervention. This model-generated data can then be used to train a robust, high-coverage state-based policy $\pi_s(a_t|s_t)$ using demonstration-bootstrapped reinforcement learning (Eq \ref{rl_finetune}). Finally, for real-world transfer from visual observations this state-based policy $\pi_s(a_t|s_t)$ is distilled into a ``fine-tuned" visuomotor policy $\pi_{Gf}(a_t|o_t)$, by collecting a set of successful rollouts $\mathcal{D}$ with $\pi_s(a_t|s_t)$ and fine-tuning the previously obtained generalist policy $\pi_G(a_t|o_t)$ as in Eq \ref{eq_distill}. This approach allows the model to retain the generalist capabilities of $\pi_G$ while achieving high success in $\mathcal{E}_{\text{test}}$. Importantly, this fine-tuning step is accomplished using only a video scan of the environment, without the need for human-provided demonstrations or feedback in the physical environment. (See Algorithm \ref{alg:scanneddeployment}, in Appendix \ref{apdx:scanneddeployment}). Finally, in the Appendix \ref{apdx:fewshotfinetuning}, we propose a second technique involving few-shot supervised fine-tuning using a limited set of human-provided demonstrations.



\section{Experimental Evaluation}
\label{sec:results}

Our experiments are designed to answer the following questions:
(a) What are the scaling laws of \MethodName? 
(b) How much can we amortize the quantity of human data needed through learning without a loss in performance? 
(c) What are the few-shot/scanned fine-tuning capabilities of the learned generalist policies?
(d) Do these scaling laws hold across different tasks?
(e) Do these generalist policies extrapolate to multi-object environments when trained with single object?

To answer these questions, we design two different tasks: \emph{placing bowls/mugs/cups in sinks} and \emph{placing boxes in shelves}. We use a single-arm manipulator, the Franka Research 3 arm with 7 DoF and a parallel jaw gripper, see Appendix \ref{apdx:hardware}. We crowdsourced environment data collection, obtaining (a maximum of ) 56 and 36 different scenes for the two tasks, respectively. We evaluated the policies across two institutions on 8 and 2 real-world scenes not included in the training set. Further details on the hardware setup and tasks are provided in Appendix \ref{apdx:tasks} and \ref{apdx:hardware}.

\subsection{Zero-Shot Scaling Laws Analysis}
\label{sec:reszeroshot}

In this section, we analyze the zero-shot performance of multiple generalist policies trained with varying amounts of training environments on the task of \emph{put a mug/bowl/cup in a sink}. For fair comparison, we train these policies using human demonstrations in each environment. In Section \ref{sec:resamortized}, we compare this baseline to the autonomous data collection system presented in Section \ref{sec:autonomousdatacollection}.  

\begin{figure}[h!]
    \centering
    \includegraphics[width=\linewidth]{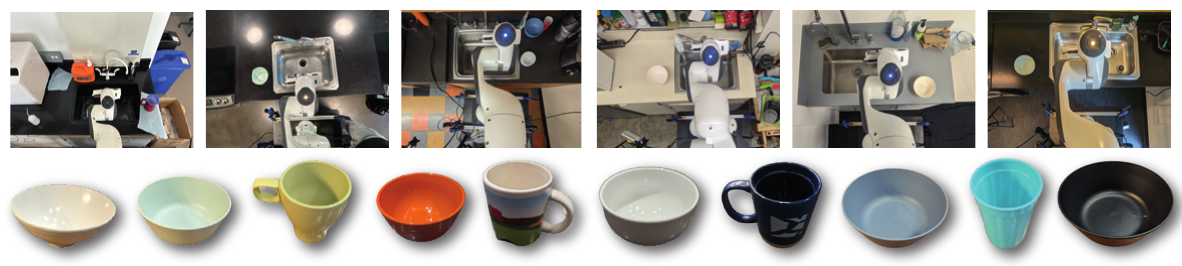}
    \caption{\footnotesize{Overview of a selected number of scenes and objects used for the real-world evaluation of the task of placing bowls/mugs/cups in the sink.} }
    \label{fig:varietyscenes}
\end{figure}

The first experiment involves a thorough real-world evaluation of these policies across two institutions, using three different kitchens and six different objects, with six rollouts each (a total of 108 rollouts per policy). As shown in Figure \ref{fig:zeroshotfig} \emph{a}, we confirm the real-to-sim-to-real pipeline scaling law: as the number of trained environments increases, the zero-shot success rate also increases, reaching a 62\% when trained on 56 environments. Furthermore, Figure \ref{fig:zeroshotfig} b shows a linear correlation between simulation and real world performance, indicating that our real-to-sim-to-real scaling approach in simulation proportionally corresponds to improved performance in the real world. 
\begin{figure}[t!]

    \centering
    \includegraphics[width=\linewidth]{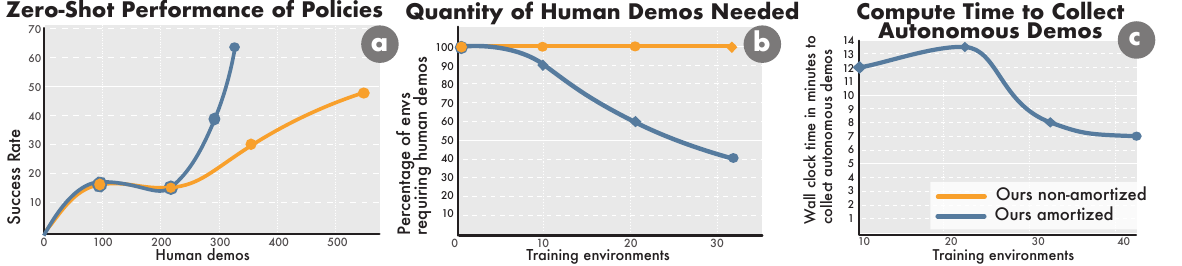}
    \caption{\footnotesize{\emph{a}) \MethodName with continual data collection becomes more efficient in number of human demos and achieves higher performance than running \MethodName uniquely from human demos. \emph{b}) with continual data collection the number of human demos required decreases throughout training. \emph{c}) even though \MethodName relies on compute we observe the amount of compute needed also tends to decrease when scaling up this process.}}
    \label{fig:continuallearningresults}
\end{figure} 
\begin{figure*}[t]
    \centering
    \includegraphics[width=\linewidth]{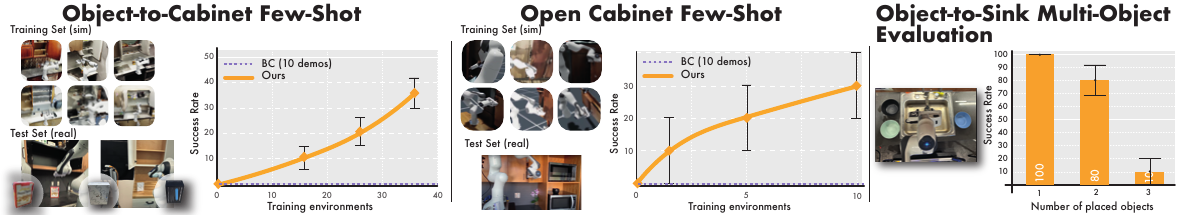}
    \caption{\footnotesize{\emph{left}: results for few-shot fine-tuning on the task of \emph{pick and place a box on a shelf} \emph{middle}: results opening a cabinet \emph{right}: multi-object evaluation results on the task of \emph{pick and place mug/bowl/cups in the sink}} }
    \label{fig:scalinglaws}
    \vspace{-10pt}
\end{figure*}
To verify the robustness of the learned policies, we ran evaluation on eight additional kitchens. The results highlight an improvement of 16\% to 60\% rate as the number of training environments increased from 9 to 56 (Figure \ref{fig:zeroshotfig} c). Figure \ref{fig:varietyscenes} shows a sample of the objects and environments used for evaluation. Finally, we stress-tested against other types of robustness (Figure \ref{fig:zeroshotfig}), including extreme lighting changes, clutter and physical disturbances, and observed that the policies suffer a drop in performance but keep obtaining success rates above 30\% (see Appendix \ref{apdx:robustness}). On the same lines we evaluate the policy on multiple objects in the scene and observe that even though it was only trained to pick up one object, it still succeeds 10\% of the times to clean a scene with 3 objects (See Figure \ref{fig:scalinglaws} and Appendix \ref{apdx:multiobject}). 

\subsection{Amortized Human Data Needed Through Continual Data Collection}
\label{sec:resamortized}

\begin{figure}[b]
    \centering
    \includegraphics[width=\linewidth]{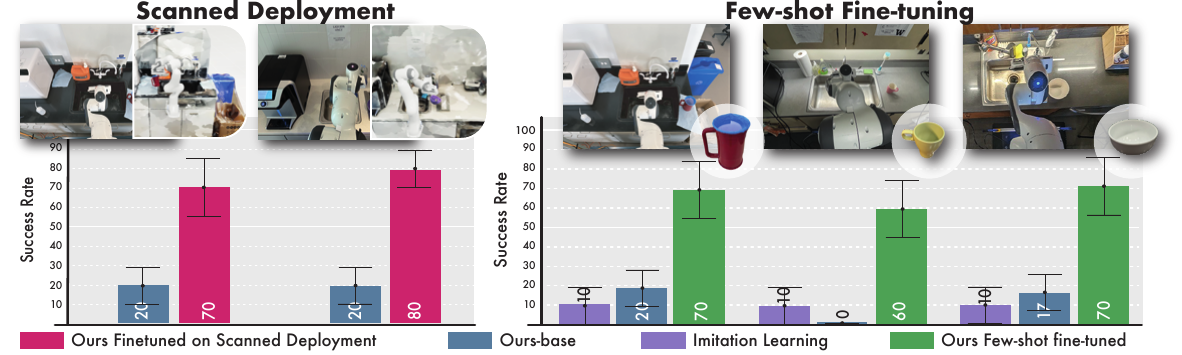}
\caption{\footnotesize{Fine-tuning results. \emph{left}: \MethodName successfully improves its performance fine-tuning autonomously on a scanned deployment environment. \emph{right}: Few-shot fine-tuning on 10 demos we can significantly improve the performance of the generalist policy on the target scene.} }
    \label{fig:finetuningresults}
\end{figure}

In this section, we evaluate the amortization of number of human demonstrations needed as learning progresses across multiple environments. We compare two approaches: our proposed system using continual data collection performed in four sequential batches of 10 environments each, and another baseline providing human demonstrations for each environment individually. The evaluation is conducted in a single real-world kitchen with six different objects for the task of \emph{put a bowl/mug/cup in a sink}, performing 6 rollouts per object. Figure \ref{fig:continuallearningresults}(a) shows that the performance per number of demonstrations significantly increases as the policy starts developing generalization. Specifically, as shown in Figure \ref{fig:continuallearningresults}(b), the quantity of human demonstrations needed decreases as the policy improves with each subsequent batch. Although \MethodName shifts the burden to compute rather than human effort, Figure \ref{fig:continuallearningresults}(c) indicates that the compute required decreases as well when scaling up the system, since the success rate of the generalist policy is higher, the number of trials performed to reach the same number of successful rollout decreases. Finally, we observe that the performance of the continually learned policy is higher than of the policy learned solely from human demonstrations. We hypothesize that this is due to the multimodality in behaviors from the human demonstrations. When the policy autonomously collects the data, behaviors remain closer to those already learned, whereas human-provided demonstrations may introduce more variability, making learning harder. 


\subsection{Fine-Tuning of Generalist Policies}
\label{sec:resfinetuning}

\textbf{Unsupervised scanned deployment fine-tuning}: To evaluate the efficacy of unsupervised fine-tuning through a scan (Section \ref{sec:finetuning}), we select two scenes for the task of \emph{placing a mug/cup/bowl in a sink} where the policy trained on 36 environments performs poorly ($\leq$ 20\%). We then apply the scanned deployment fine-tuning algorithm as described in Section \ref{sec:finetuning}. As shown in Figure \ref{fig:finetuningresults}, this results in an average performance increase of 55\% without any additional human demonstrations. 

\textbf{Few-shot supervised fine-tuning}: We select three environments where the base policy trained on 36 environments performs poorly ($\leq$20\%). We then collect 10 demonstrations for each environment and apply the few-shot fine-tuning procedure described in Section \ref{apdx:fewshotfinetuning}. This fine-tuning improves the performance of the base policy by an average of a 54\% in success rate.

\subsection{Analysis of \MethodName on More Tasks }

We attempt to solve two additional tasks, \emph{putting a box on a cabinet} and \emph{opening a cabinet}. The first is a more complicated manipulation task since it requires more precise grasping to not make the box fall, and the second shows how our proposed method works for articulated objects (see Appendix \ref{apdx:real2sim} where we give more details on how the proposed real-to-sim pipeline can handle articulated objects). In these two tasks, we focus our analysis on few-shot fine-tuning as described in Section \ref{sec:finetuning}. For \emph{putting a box on a cabinet}, we crowdsourced 36 environments, collected 10 demonstrations for each of three scenes, and reported the performance after fine-tuning with 10 demos. For \emph{opening a cabinet}, we crowdsource 10 environments, collect 10 demonstrations for a new test environment and report the performance after fine-tuning with these demos. In Figure \ref{fig:scalinglaws}, we show the performance the performance increases with the number of training environments, without reaching a saturation point. Fine-tuning the policy trained on 36 and 10 environments respectively resulted in a significant performance improvement of 36\% and 30\% compared to the imitation learning baseline, which had a 0\% success rate. We expect the performance of the generalist policies to keep improving as we have not reached a saturation point.

\section{Conclusion}
\label{sec:conclusion}
In this work, we present a system for scaling up robot learning through crowdsourced simulation. We showed that through the learning of visual generalist policies, we are able to scale across environments with decreasing amounts of human effort. The resulting policies are shown to transfer to the real world, enabling both zero-shot and finetuning results. 

\emph{Limitations}: While with this work we demonstrate superlinear scaling of data with respect to human demonstrations, the burden shifts to compute. And even though we have shown a reduction in compute time with scaling, it still exceeds the time required for collecting real-world demonstrations. Additionally, training in simulation poses challenges, as not all real-world objects can be accurately simulated yet, such as liquids and deformable objects. However, contrary to the human teleoperation efforts, with advancements in compute resources and simulator research, systems like \MethodName will benefit from these and further improve scalability. \emph{Conclusion}: This work presents \MethodName, a real-to-sim-to-real system that trains generalist policies with sublinear human effort. This research paves the way for building robotic foundation models in simulation with larger datasets and enhanced robustness.

\section{Acknowledgements}
The authors would like to thank the Improbable AI Lab and the WEIRD Lab members for their valuable feedback and support in developing this project. Marcel Torne was partly supported by the Sony Research Award, the US Government, and the Hyundai Motor Company, and Arhan Jain was supported by funding from the Army Research Lab. This work was also partially supported by the Amazon Science Hub. 

\balance

\hfill \break
\textbf{Author Contributions}

\textbf{Marcel Torne} led the project and jointly conceived the overall project goals and methods of \MethodName. Led the technical development of the method. Marcel helped run some of the real-world experiments. Marcel wrote the paper and made the main figures of the paper. Marcel also led the design of the website.

\textbf{Arhan Jain} jointly conceived the overall project goals and methods of \MethodName. Made major technical contributions to the implementation of the method. Was responsible for part of the real-world experiments. Arhan made main contributions to the website, he is the author of the interactive web viewer. 

\textbf{Jiayi Yuan} led the large-scale training on the crowdsourced environments. Jiayi also helped with building the website, some figures in the paper, part of the real-world experiments, and writing the appendix.

\textbf{Vidaaranya Macha} led the real-world experiments and data collection.

\textbf{Lars Ankile} was involved in regular discussions about implementation challenges and when conceiving the paper.

\textbf{Anthony Simeonov} was involved in conceiving the goals and motivation of the project and provided crucial advice throughout the project.

\textbf{Pulkit Agrawal} was involved in conceiving the project's goals, suggesting baselines and ablations, and helped edit the paper.

\textbf{Abhishek Gupta} was involved in conceiving the project's goals, suggested baselines and ablations, helped edit the paper, and was the main advisor of the project.


\clearpage


\bibliography{references}  

\clearpage
\section*{Appendix}
In the Appendix, we will cover the following details of our work.
\begin{itemize}
    \item \textbf{Method Details} Appendix \ref{apdx:method}: Details about amortized data collection algorithms, real-to-sim transfer, and autonomous data collection used for fine-tuning and teacher-student distillation.
    \item \textbf{Task Details} Appendix \ref{apdx:tasks}: Details of tasks used for evaluating \MethodName.
    \item \textbf{Implementation Details} Appendix \ref{apdx:implementation}: Specification of hyper-parameters used in the network architectures, point-cloud processing, and dataset used in \MethodName.
    \item \textbf{Detailed Evaluation Result} Appendix \ref{apdx:evaluation}: Detailed results of the evaluation, including robustness experiments, adding disturbance and distractors.
    \item \textbf{Hardware Setup} Appendix \ref{apdx:hardware}: Specification for hardware setup used for training and evaluating \MethodName.
    \item \textbf{Crowdsourcing} Appendix \ref{apdx:crowdsourcing}: Specification for crowdsourcing real-world 3D scans.
    \item \textbf{Compute Resources} Appendix \ref{apdx:compute}: Specifications for compute resources used for data collection, training, and evaluating \MethodName.
\end{itemize}

\section{Method Details}
\label{apdx:method}

\subsection{Amortized Data Collection}

We present the following algorithm to explain the amortized data collection section in \MethodName.
\begin{algorithm}[h!]
\begin{algorithmic}[1]
\STATE{}\textbf{Input:} Human demonstrator $\mathcal{H}$, crowdsource humans $\mathcal{C}$
\STATE{} Initialize vision-based generalist policy $\pi_G$
\WHILE{True}
\STATE Sample set of $K$ digital twins from crowdsourced humans $\{\mathcal{E}_K, \mathcal{E}_{K+1}, \dots, \mathcal{E}_{2K} \} \sim \mathcal{C}$
\STATE $\mathcal{T} \leftarrow \{\}$
\FOR{ $\mathcal{E}_i$ in $\mathcal{E}_K, \mathcal{E}_{K+1}, \dots, \mathcal{E}_{2K}$}
\STATE $\mathcal{T}_e \leftarrow \mathrm{RolloutPolicy}(\mathcal{E}_i, \pi_G) $
\STATE $\mathcal{T} \leftarrow \mathcal{T} \cup \mathrm{FilterSuccessfulRollouts}(\mathcal{T}_e) $
\ENDFOR{}
\STATE $\pi_s \leftarrow \mathrm{RLFinetuning}(\mathcal{T}, \{\mathcal{E}_K, \mathcal{E}_{K+1}, \dots, \mathcal{E}_{2K} \})$
\STATE $\mathcal{T}_h \leftarrow \{\}$
\STATE $\mathcal{F} \leftarrow  \mathrm{FailedEnvironments}(\{\mathcal{E}_K, \mathcal{E}_{K+1}, \dots, \mathcal{E}_{2K} \}, \pi_s)$
\FOR{ $\mathcal{E}_i$ in $\mathcal{F}$}
\STATE $\mathcal{T}_h \leftarrow \mathcal{T}_h \cup \mathrm{CollectDemos}(\mathcal{E}_i, \mathcal{H}) $
\ENDFOR{}
\STATE $\pi_h \leftarrow \mathrm{PPORLFinetuning}(\mathcal{F}, \pi_h)$
\STATE $\pi_G \leftarrow \mathrm{TeacherStudentDistillation}(\mathcal{E}, \pi_G, \pi_s, \pi_h)$
\ENDWHILE
\end{algorithmic}
  \caption{\MethodName: Amortized Data Collection for Generalist Policies} 
  \label{alg:amortized}
\end{algorithm}

\subsection{Real-to-Sim Transfer of Scenes}
\label{apdx:real2sim}

\begin{figure}[h!]
    \centering
    \includegraphics[width=\linewidth]{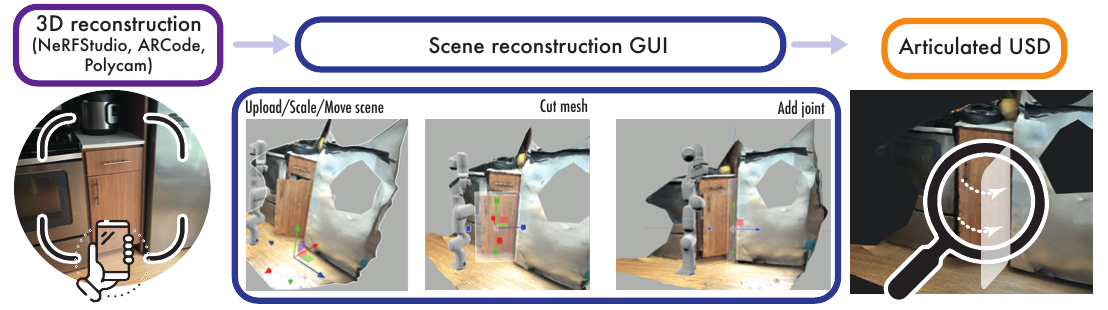}
    \caption{\footnotesize{Overview of the GUI used for this project, it allows to extract a 3D mesh from a video, articulate the scene and export it in USD format.} }
    \label{fig:real2simscenes}
\end{figure}

Unlike prior work~\citep{torne2024reconciling, chen2024urdformer}, our goal is not to accurately master a single environment, but rather to train a generalist agent capable of generalizing to new, unseen environments. To obtain a wide distribution of scenes with a variety of layouts, colors, and lighting conditions, We developed our general purpose, an easy-to-use, real-to-sim pipeline that supports the crowdsourcing contribution of 3D scans (See Figure \ref{fig:real2simscenes} for an overview of the GUI \citep{torne2024reconciling}). Digital twins are obtained directly from real-world videos or image sequences using photogrammetry methods such as Gaussian splatting  \citep{kerbl3Dgaussians} and neural radiance fields \citep{nerf}. High-fidelity 3D meshes can be scanned in under five minutes using off-the-shelf mobile software such as Polycam \citep{polycam2020} and ARCode \citep{arcode2022}. In Table \ref{tab:timereal2sim}, we show the low average time needed to create a scene with different configurations. This easy-to-use software running on standard, commercial mobile phones enables crowdsourcing of real-world scans from non-experts worldwide with minimal instruction. The crowdsourced scenes demonstrate a natural distribution of clutter, scene layouts, colors, lighting conditions, and positional variations. 

\begin{table*}
\centering
\begin{tabularx}{\textwidth}{@{}lc*{6}{>{\centering\arraybackslash}X}@{}}
\toprule
 Task  & Scanning Scene & Scanning Objects  & Setting Up Object-to-Sink Scene  & Setting Up Object-to-Cabinet Scene  & Setting Up Open Cabinet Scene (Articulated)   \\
 \midrule
\addlinespace
  Time & 3 min 15 sec & 4 min 50 sec  & 1 min 30 sec & 1 min 30 sec & 3 min \\
 
\bottomrule
\addlinespace

\end{tabularx}
\caption{Average time needed to set up different kinds of scenes with the proposed Real-to-Sim pipeline.}
\label{tab:timereal2sim}
\end{table*}
				
These real-world scans are then easily transferred into a photo-realistic physics simulator, Issac Sim \citep{isaacsim2022}, using an easy-to-use GUI for scene articulation and curation \citep{torne2024reconciling}.  This flexible interface accommodates various scene complexities, from static to highly articulated environments. Using the GUI, we also add objects of interest (bowl/mug/cup for putting the object into the sink, box for putting the box in the cabinet) into the scene, and additional sites to mark the position of the sink and cabinet.

With the scene and the object rendered inside the simulation, we use teleoperation with a keyboard to collect 10 demonstrations for each articulated environment. There are 14 different discretized actions to choose from, corresponding to two directions in all spatial axes and rotational axes, and open and close the gripper. See Appendix \ref{apdx:tasks} for details.

\subsection{Autonomous Data Collection}
\label{apdx:continuallearning}

\subsubsection{Multi-task bootstrapped RL fine-tuning:}
\label{apdx:multitaskrl}

Given a set $\mathcal{T}$ of 10 demonstrations on each one of the digital twin scenes in the batch, and an easily defined sparse reward across all tasks, we leverage the current capabilities of fast multi-environment training on GPUs and accurate simulators to do RL fine-tuning using PPO \citep{schulman2017proximal} to obtain a policy that is more robust to object poses, corrections, and disturbances than if we simply learned from the demos. In addition, this multi-scene policy is being trained from privileged state space, since this removes the need for rendering, making the amount of simulated parallel environments higher, and training becomes faster since we can use bigger batch sizes. Finally, we observe that although equivalent in theory, training across multiple scenes instead of a single environment at a time per GPU in practice brings a big speedup in training. With the available resources (training on a 3090 RTX NVIDIA GPU) our experiments are run with 10 different scenes in parallel spread across 2048 environments, and even though the total training time is the same as for a single scene, the policy now works across 10 scenes which corresponds effectively to a 10x speedup. 

\subsubsection{Teacher-student distillation} 
\label{apdx:teacherstudentdistill}

In the previous RL fine-tuning step, we obtained a state-based policy that works across the whole batch of environments. However, in the real world, we do not have access to this privileged state of the environment such as object poses. For this reason, we need our policy to take as input a state representation available in the real world. We decide to use colored point clouds as the sensor modality. Thereafter, we use teacher-student distillation techniques to distill the obtained policies into $\pi_G$. This consists of for each one of the scenes we use the working state-based policy to collect a set of 1000 trajectories. Out of the 1000 trajectories, 500 of them are rendered from two cameras in simulation and 500 are synthetically generated by sampling from the meshes, making the point cloud fully observable. In practice, the synthetically generated point clouds make learning with point clouds as input much smoother even for the camera-rendered point clouds. Due to the significant amount of data available, our experiments go up to 56 environments with 1000 trajectories each, there are some important implementation details for the point cloud policy that make this learning feasible. The size of the dataset is 56000 trajectories with on average 120 steps per trajectory, making the total dataset size to be 6.720.000 state-action pairs. This dataset of point clouds does not fit into memory, thereafter, we need to employ different techniques to keep balanced batches. We store all of the data in disk and for every batch, we query around 5 trajectories per environment and load them all in memory. Then we iterate over this new batch in mini-batches as big as the GPU can hold and keep accumulating gradients to avoid the GPU running out of memory. Finally, additional details are explained in Appendix \ref{apdx:pcdpolicy}. We avoid catastrophic forgetting by retaining the data from previous batches during distillation. 

\subsection{Fine-tuning}
\label{apdx:finetuning}
\subsubsection{Scanned Deployment Fine-tuning}
\label{apdx:scanneddeployment}
\begin{algorithm}[h!]
\begin{algorithmic}[1]
\STATE{}\textbf{Input:} a generalist policy $\pi_{G}$, a digital twin of an environment $\mathcal{E}$
\STATE $\mathcal{T} \leftarrow \{\}$
\WHILE{$\vert \mathcal{T} \vert \leq$  10}  
\STATE $\mathcal{T}_e \leftarrow \mathrm{RolloutPolicy}(\mathcal{E}, \pi_G) $
\STATE $\mathcal{T} \leftarrow \mathcal{T} \cup \mathrm{FilterSuccessfulRollouts}(\mathcal{T}_e) $
\STATE $\pi_s \leftarrow \mathrm{RLFinetuning}(\mathcal{T}, \mathcal{E})$
\STATE $\pi_G \leftarrow \mathrm{TeacherStudentDist}(\mathcal{E}, \mathrm{FreezeEncoder}(\pi_G), \pi_s)$
\ENDWHILE
\end{algorithmic}
  \caption{Scanned-deployment fine-tuning} 
  \label{alg:scanneddeployment}
\end{algorithm}

\subsubsection{Few-shot Supervised Fine-tuning:} 
\label{apdx:fewshotfinetuning}


The second proposed fine-tuning technique involves using small amounts of human-provided real-world demonstrations for few-shot supervised fine-tuning. We fine-tune the generalist policy $\pi_G(a_t|o_t)$ using supervised learning on a dataset of human-collected visuomotor demonstrations $\mathcal{D}_h$ via standard maximum likelihood as shown in Eq \ref{eq_distill}.
 
Architecturally, this involves freezing the preliminary ``visual processing" layers and fine-tuning only the final fully-connected layers of the pretrained generalist network $\pi_G$. As we show in Section \ref{sec:resfinetuning}, this straightforward fine-tuning procedure can yield significant performance improvements with a small number of real-world demonstrations.

\section{Task Details}
\label{apdx:tasks}

In this section of the appendix, we describe the specification of each task for training and evaluating \MethodName. For each task, the state space consists of the concatenation of the following state information: object positions, object orientations, DOF positions of the tool normalized to its max and min ranges, end-effector orientation, and end-effector position. The action space consists of one of the 14 discretized actions, corresponding to the end effector delta pose. In specific, the 14 actions include the following: 6 actions in position change, corresponding to $\pm 0.03$m change in each axis; 6 actions in orientation change, corresponding to $\pm 0.02$ radian change in each axis; 2 actions corresponding to gripper open and close.

We define a sparse reward function for each task in PPO training:
\begin{itemize}
    \item \textbf{Put object into sink}: $ success =$ \\$||\text{sink\_site}-\text{object\_site}||_2<0.25 \\ ~\&\&~  \text{condition(object\_upright)} ~\&\&~ \text{condition(gripper\_open)}$
    \item \textbf{Put object into cabinet}:  $ success =$ \\$\text{cabinet\_z\_axis} < \text{object\_z\_axis}~\&\&~  \text{condition(object\_upright)} \\ ~\&\&~ \text{condition(gripper\_open)}$
    \item \textbf{Open cabinet}: $ success =$
    \\ $\text{cabinet\_joint} > 0.65 ~\&\&~ \text{condition(gripper\_open)} $
\end{itemize}

\subsection{Simulation details}
\label{apdx:simulationdetails}

We used the latest physics-accurate with photorealistic rendering simulation platform, Isaac Sim \citep{isaacsim2022} for our simulation task training. Our environment codebase is inspired by the Orbit codebase \citep{mittal2023orbit}, a unified and modular framework for robot learning built upon Isaac Sim.

\begin{figure}[t!]
    \centering
    \includegraphics[width=\linewidth]{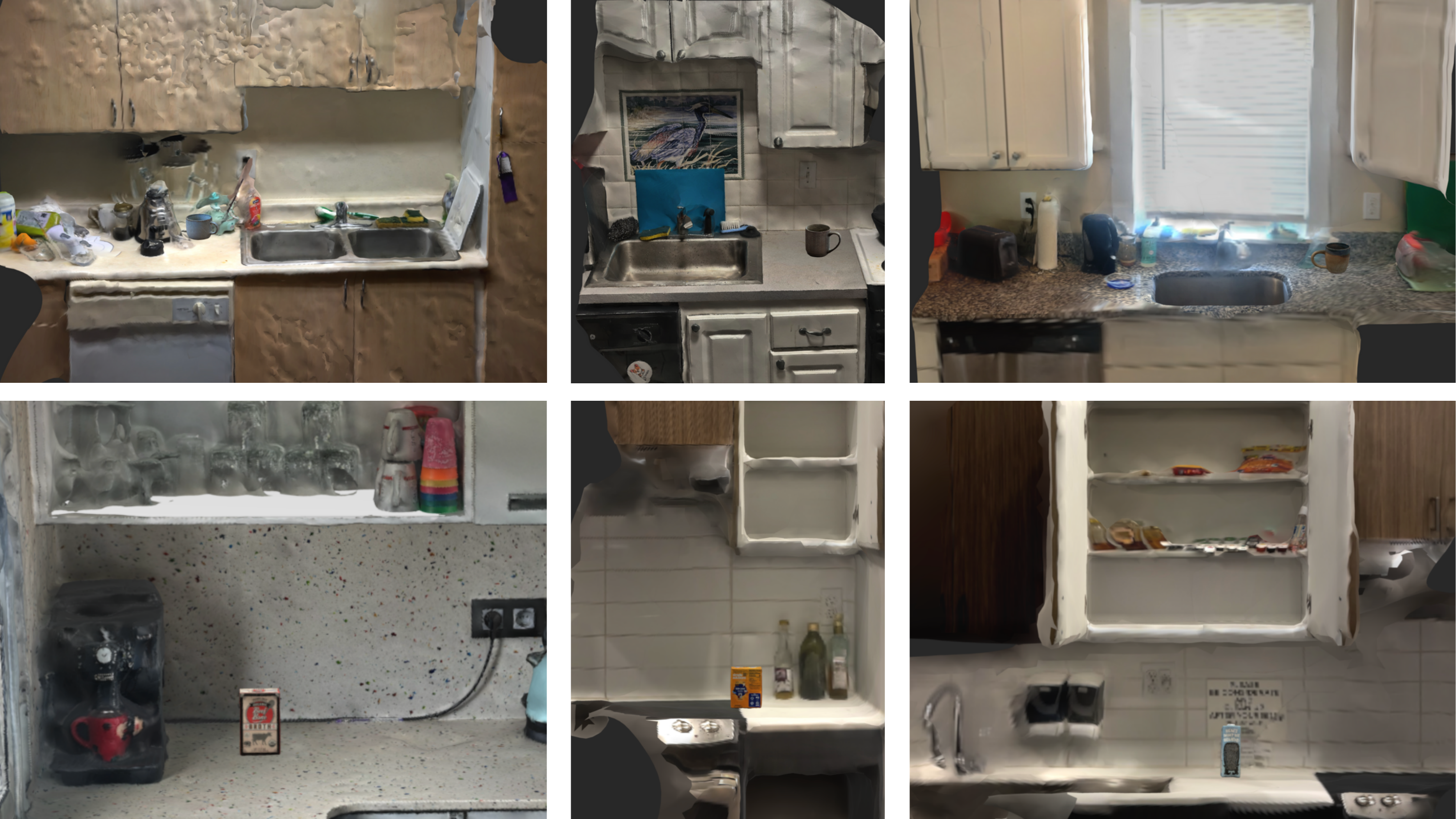}
    \caption{\footnotesize{Examples of simulated environment used for RL fine-tuning. The top ones correspond to environments for the obj2sink task and the bottom ones correspond to environments for the obj2cabinet task.} }
    \label{fig:simulation_examples}
\end{figure}

For the simulation parameters of the environment, we use the default value set by the GUI, except we change the collision mesh of the scene from convex decomposition to SDF mesh with 256 resolution to reflect high-fidelity collision mesh. For all the other objects, we use the default value, which is convex decomposition with 64 hull vertices and 32 convex hulls as the collision mesh for all objects. We keep all the physical parameters of the environment as default in the GUI. The default value of physical parameters for all objects are as follows: dynamic and static frictions of all objects as 0.5, the joint frictions as 0.1, and the mass as 0.41kg. See Table \ref{tab:random-paramstask} for task-specific randomization parameters, Table \ref{tab:cam-paramstask} for task-specific camera parameters, and Figure \ref{fig:simulation_examples} for examples of simulated scenes.

\begin{table*}
\centering
\begin{tabularx}{\textwidth}{@{}lc*{6}{>{\centering\arraybackslash}X}@{}}
\toprule
 Task  & Episode & Randomized  & Position  & Position  & Orientation & Orientation    \\
  & length & Object Ids &  Min (x,y,z) &  Max (x,y,z) & Min (z-axis) & Max (z-axis)  \\
 \midrule
\addlinespace
  obj2sink & 135 & [Background, Object] & [[-0.1,-0.1,-0.1], [-0.1,-0.1,0]]  & [[0.1,0.1,0.1], [0.1,0.1,0]]  & [-0.3, -0.3] & [0.3, 0.3]\\
 obj2cabinet & 150 & [Background, Object]  & [[-0.1,-0.1,-0.05], [-0.1,-0.1,0]] & [[0.1,0.1,0.05], [0.1,0.1,0]]  & [-0.1, -0.15] & [0.1, 0.15]\\
 
\midrule
\addlinespace

\bottomrule
\end{tabularx}
\vspace{0.05cm}
\caption{Specific simulation parameters for each task.}
\label{tab:random-paramstask}
\end{table*}

\begin{table*}
\centering
\begin{tabularx}{\textwidth}{@{}lc*{6}{>{\centering\arraybackslash}X}@{}}
\toprule
 Task & Position (x,y,z) & Rotation (quat) & Crop Min & Crop Max & Size    \\
  Parameters & Camera   & Camera  & Camera  & Camera  & Image \\
 \midrule
\addlinespace
 obj2sink & [-0.01, -0.50, 0.69],& [0.84,0.33, -0.15, -0.41],& [-0.8,-0.8,-0.3] & [0.8, 0.8, 1.0] & (640,480)\\
 &[-0.01, -0.50, 0.69]&[-0.42,-0.22,-0.39,0.79]&& \\
 obj2cabinet & [-0.01, -0.50, 0.69],& [0.84, 0.33, -0.15, -0.41],& [-0.2, -0.5, -0.5] & [1.5, 0.5, 1.5] & (640,480) \\
  &[-0.01, -0.50, 0.69]&[-0.42, -0.22, -0.39, 0.79]&& \\
\midrule
\addlinespace

\bottomrule
\end{tabularx}
\vspace{0.05cm}
\caption{Camera parameters for each task.}
\label{tab:cam-paramstask}
\end{table*}

\section{Implementation Details}
\label{apdx:implementation}

\subsection{Architecture Details}
\subsubsection{State-based policy}
\label{apdx:statebasedpolicy}
As described in Section \ref{sec:autonomousdatacollection}, we trained a series of state-based policies with privileged information in simulation. The policy model is a simple Multi-Layer Perceptron (MLP) network, with input as the privileged state in simulation as specified in \ref{apdx:tasks} and outputs a probability distribution of 14 classes, corresponding to the probabilities for each discrete end-effector action. To implement PPO with the BC loss algorithm, we built upon the Stable Baselines 3 repository \citep{stable-baselines3}. The size of the MLP network is a mix of two sizes: two layers of size 256 and 256, and three layers of size 256, 512, and 256. See Table \ref{tab:ppoparams} for more details.
\begin{table*}
\centering
\begin{tabularx}{\textwidth}{@{}lc*{6}{>{\centering\arraybackslash}X}@{}}
\toprule
  MLP layers & PPO n\_steps & PPO batch size & PPO BC batch size & PPO BC weight & Gradient Clipping\\
\midrule
\addlinespace
 256,256 or 256, 512, 256 & Episode length &  31257 & 32 & 0.1 & 5\\

\bottomrule
\end{tabularx}
\vspace{0.05cm}
\caption{State-based policy training parameters. The rest of the parameters are the default as described in Stable Baselines 3\citep{stable-baselines3}. }
\label{tab:ppoparams}
\end{table*}

\subsubsection{Point cloud policy}
\label{apdx:pcdpolicy}

As mentioned in Section \ref{sec:finetuning}, when distilling the state-based teacher policy to a fine-tuned visuomotor policy, we will train a point cloud policy as the student. We train an MLP network of size 256,256, that takes the embedding of the point cloud observation, which has 128 dimensions, together with the state of the robot (end-effector scaled pose, position, and orientation), that has 9 dimensions, as the input, and a probability distribution of 14 actions as output. To encode the point cloud observation, we use the volumetric 3D point cloud encoder proposed in Convolutional Occupancy Networks \citep{peng2020convolutional}, which consists of a local point net that converts the point cloud to 3D features, followed by a 3D U-Net that output a dense voxel of features. The output features are then pooled by a max pooling layer and an average pooling layer separately, with the pooling output concatenated into the resulting encoding of dimension 128. We base our code in \citep{torne2024reconciling}.

\section{Baseline Comparisons with Large Scale Real World Data Collection}

We deployed two of the state-of-the-art robot models trained on the Open X-Embodiement dataset \citep{padalkar2023open} consisting of more than 800k real-world trajectories: OpenVLA\citep{kim2024openvla}, a vision-language-action model consisting of fine-tuned Llama 2 7B fine-tuned and Octo\citep{team2024octo}, a pre-trained transformer based diffusion policy.

We provided RGB images and language instructions as inputs to the model and evaluated the model zero-shot on the task of moving an object to sink. While neither of the two models successfully completes the task, OpenVLA performs qualitatively better than Octo (see the additional material videos).

We then collected 10 demos and fine-tuned the Octo policy with these. However, we observe that the performance for finetuned Octo still yields zero success, with qualitatively better performance than the zero-shot result.

These results, presented in Table \ref{tab:baselines}, show that zero-shot and few-shot performance on the tasks that we are tackling in this paper is not solved yet. It also shows the need for much more data to be collected in order to solve these tasks in a variety of scenarios. Thereafter providing evidence of the benefits of \MethodName and of leveraging data from simulation to scale up the robot data collection.

\begin{table*}
\centering
\begin{tabularx}{\textwidth}{@{}lc*{6}{>{\centering\arraybackslash}X}@{}}
\toprule
 Method  & \MethodName & Imitation Learning (point clouds)  & OpenVLA (zero-shot RGB)  & Octo (zero-shot RGB)  & Octo (fine-tuned RGB)    \\
 \midrule
\addlinespace
  Success Rate & 62 $\pm$ 5 & $10 \pm 5$  & $0 \pm 0$ & $0 \pm 0$ & 0 $\pm$ 0 \\
 
\midrule
\addlinespace

\bottomrule
\end{tabularx}
\vspace{0.05cm}

\caption{Baseline comparison with large scale real-world data collection policies.}
\label{tab:baselines}
\end{table*}

\section{Detailed Evaluation Results}
\label{apdx:evaluation}
We conducted experiments involving disturbance and distractors for putting the object into the sink task to study the robustness of the generalist policies. The experiments include multi-object scenarios, dim lighting scenarios, messy kitchen scenarios, and disturbance scenarios.

\subsection{Evaluation on Multi-Object Scenes}
\label{apdx:multiobject}
In this section, we study the extrapolation and robustness capabilities of the learned generalist policies by evaluating them on tasks involving multi-object scenes. Specifically, the robot needs to pick and place multiple objects into the sink sequentially, even though it was trained on single objects. We evaluated this by allowing the robot six trials to place the three objects in the kitchen into the sink. As shown in Figure \ref{fig:scalinglaws}, despite not being trained for multi-object, the policy succeeds 80\% of the time in placing two objects and 10\% in placing all three objects sequentially. 

\subsection{Evaluation on Scenes Involving Disturbance and Distractors }
\label{apdx:robustness} 

\begin{figure}[t!]
    \centering
    \includegraphics[width=\linewidth]{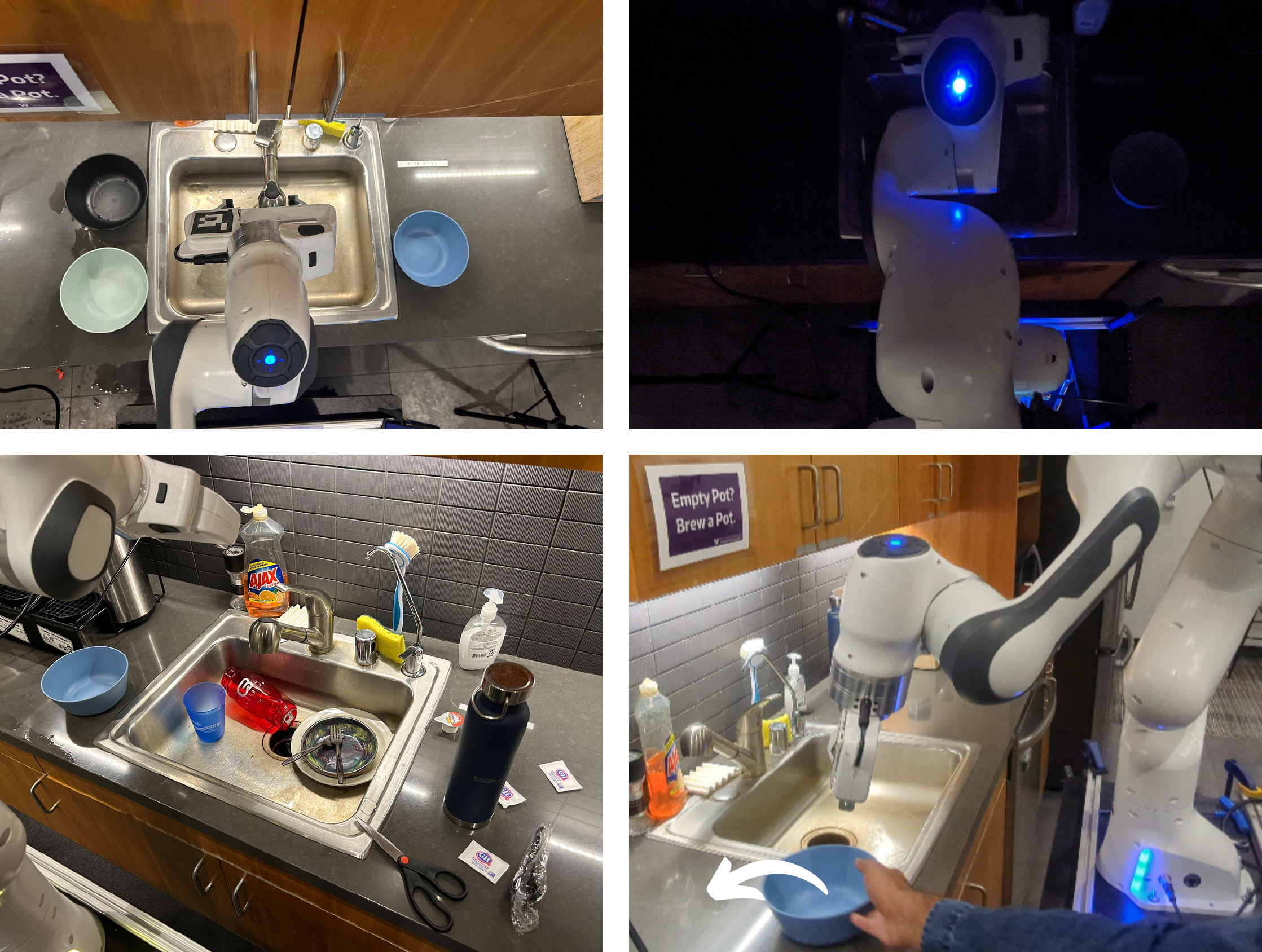}
    \caption{\footnotesize{Overview of the experiment setup for evaluating the robustness capacity of the generalist policy. upper left: multi-object scenario. upper right: dim lighting scenario. bottom left: messy kitchen scenario. bottom right: human disturbance scenario. See Table \ref{tab:robustness_2} for success rate.  } }
    \label{fig:robustness_setup}
\end{figure}

In the dim lighting scenario, there is minimal lighting in the scene, while the robot is only trained in the environment with sufficient lighting. The robot was able to complete the task successfully into the sink for $30\%$ of all trials. See Figure \ref{fig:robustness_setup} for the experimental setup.

In the messy kitchen scenario, dirty dishes and tableware are sitting in the sink, closely mimicking the realistic setting of a household kitchen sink. The robot is only trained in an environment with a clean sink. The robot was able to complete the task successfully into the sink for $30\%$ of all trials. See Figure \ref{fig:robustness_setup} for the experimental setup.

In the human disturbance scenario, the experimenter pushes the object the change its position during the evaluation process. The robot is able to complete the task successfully into the sink for $50\%$ of all trials. See Figure \ref{fig:robustness_setup} for the experimental setup.



\begin{table*}
\centering
\begin{tabularx}{\textwidth}{@{}lc*{7}{>{\centering\arraybackslash}X}@{}}
\toprule
$\#$ of Env distilled & Kitchen Ids & Bowl right of the sink & Bowl left of the sink & Mug right of the sink & Mug left of the sink & Overall\\
 \midrule
\addlinespace
9 &  Kitchen 1 & 66.7$\%$ & 0$\%$  & 22.2$\%$  & 0$\%$ & 22.2$\%$ \\
9 &  Kitchen 2 & 66.7$\%$ & 0$\%$ & 11.1$\%$ & 0$\%$ & 19.4$\%$ \\
9 &  Kitchen 3 & 41.7$\%$ & 0$\%$ & 16.7$\%$ & 0$\%$ & 14.6$\%$\\
36 &  Kitchen 1 & 55.6$\%$ & 55.6$\%$ & 22.2$\%$ & 22.2$\%$  &  38.9$\%$\\
36 &  Kitchen 2 & 44.4$\%$ & 33.3$\%$ & 22.2$\%$ & 22.2$\%$  &  30.6$\%$\\

36 &  Kitchen 3 & 13.3$\%$ & 46.7$\%$ & 20.0$\%$  &60.0$\%$   & 35.0$\%$  \\
56 &  Kitchen 1 & 55.6$\%$  & 55.6$\%$ & 44.4$\%$& 66.7$\%$ & 64.8$\%$\\
56 &  Kitchen 2 & 77.8$\%$ & 33.3$\%$  & 55.6$\%$  & 22.2$\%$  & 47.2$\%$\\

56 &  Kitchen 3 & 58.3$\%$  & 83.3$\%$ & 66.7$\%$ & 91.7$\%$ & 75.0$\%$\\

\midrule
\addlinespace

\bottomrule
\end{tabularx}
\vspace{0.05cm}
\caption{Zero-shot success rate for putting an object to sink task. We tested different types of objects such as bowls, mugs, and cups. We evaluated the policy by placing the object on either side of the sink.}
\label{tab:sink-zero-shot}
\end{table*}

\begin{table*}
\centering
\begin{tabularx}{\textwidth}{@{}lc*{7}{>{\centering\arraybackslash}X}@{}}
\toprule
Num. Env distilled & Num. demos & Kitchen Ids & Object type & Success rate\\
 \midrule
\addlinespace
 0 (IL) & 10 &  Kitchen 1 & Bowl  & 10.0$\%$ \\
 36 & 10 &  Kitchen 1 & Bowl & 70.0$\%$ \\
 0 (IL) & 10 &  Kitchen 2 & Mug  & 10.0$\%$ \\
36 & 10 &  Kitchen 2 & Mug & 70.0$\%$ \\
 0 (IL) & 10 &  Kitchen 3  & Mug & 10.0$\%$ \\
 36 &  10 & Kitchen 3 &Mug  & 60.0$\%$ \\
 0 (IL) & 10 &  Kitchen 3 & Bowl & 10.0$\%$ \\
 36 &  10 & Kitchen 3 &Bowl & 60.0$\%$ \\

\midrule
\addlinespace

\bottomrule
\end{tabularx}
\vspace{0.05cm}
\caption{Imitation learning baseline and few-shot supervised fine-tuning success rate for putting an object to sink task.}
\label{tab:il-baseline}
\end{table*}

\begin{table*}
\centering
\begin{tabularx}{\textwidth}{@{}lc*{7}{>{\centering\arraybackslash}X}@{}}
\toprule
Num. Env distilled & Num. demos & Kitchen Id & Success rate for grasping & Success rate for placing\\
 \midrule
\addlinespace
 0 (IL) & 10 &  Kitchen 1& 0.0$\%$ & 0.0$\%$ \\
 16 & 10 &    Kitchen 1&  10.0$\%$ &  0.0$\%$ \\
 26 & 10 &   Kitchen 1& 60.0$\%$ &  20.0$\%$ \\
36 & 10 &    Kitchen 1& 80.0$\%$ &  30.0$\%$ \\
 0 (IL) & 10 &  Kitchen 2& 0.0$\%$ & 0.0$\%$ \\
 16 & 10 &    Kitchen 2&  10.0$\%$ &  0.0$\%$ \\
 26 & 10 &   Kitchen 2& 20.0$\%$ &  20.0$\%$ \\
36 & 10 &    Kitchen 2& 30.0$\%$ &  30.0$\%$ \\
\midrule
\addlinespace

\bottomrule
\end{tabularx}
\vspace{0.05cm}
\caption{Imitation learning baseline and few-shot supervised fine-tuning success rate for putting an object to cabinet task. We recorded both the success rate for grasping the object and placing the object in the cabinet.}
\label{tab:cabinet-finetune}
\end{table*}

\begin{table*}
\centering
\begin{tabularx}{\textwidth}{@{}lc*{4}{>{\centering\arraybackslash}X}@{}}
\toprule
Num. objects successfully & Generalist policy  & Imitation Learning  & Average  \\
placed in the sink & success rate & baseline success rate & num. of episodes \\
 \midrule
\addlinespace
1 & 100.0$\%$ & 0.0$\%$ & 1.6 \\
2 & 80.0$\%$  & 0.0$\%$ & 3.5 \\
 3 & 10.0$\%$ & 0.0$\%$ & 4 \\
\addlinespace

\bottomrule
\end{tabularx}
\vspace{0.05cm}
\caption{Multi-object scenario evaluation. As shown in Fig \ref{tab:robustness_2}, three objects are placed in the scene. The policy is rolled out for 6 episodes in total. The table shows the success rate for the average number of episodes it takes to place a certain number of objects into the sink.}
\label{tab:multiobjects}
\end{table*}

\section{Hardware Setup}
\label{apdx:hardware}
Real-world experiments are run on two different Panda Franka arms. Both of the Panda Franka arms are mounted on mobile tables, and run the same experiments, but they are located in two different institutions and therefore have access to different real-world kitchen settings.

\begin{figure}[t!]
    \centering
    \includegraphics[width=\linewidth]{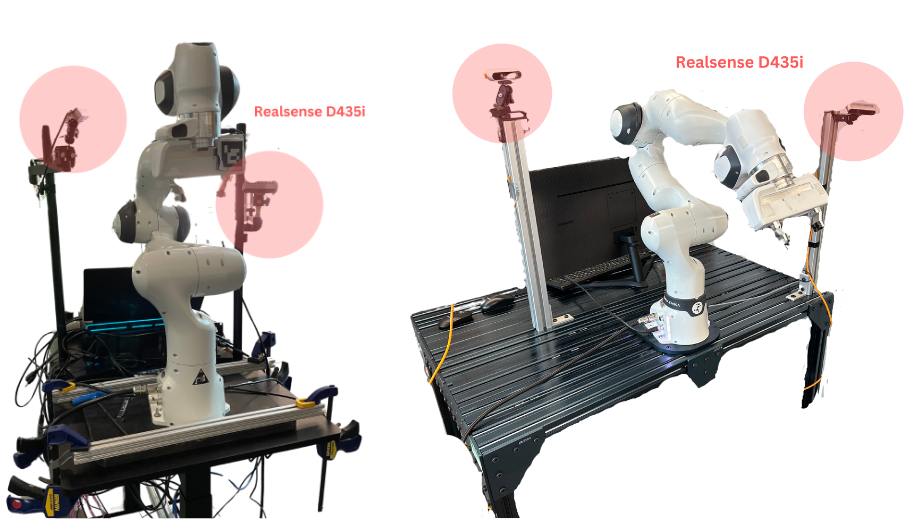}
    \label{fig:hardware}
    \caption{\footnotesize{Overview of the hardware used to evaluting \MethodName. left: used to evaluate in Kitchen 2 in both tasks. right: used to evaluate in Kitchen 1 and 3 in putting objects to sink and Kitchen 1 in putting box to cabinet.} }
    
\end{figure}

We mount two calibrated cameras per setup to obtain depth perception to create an aligned point cloud map for vision-based policies. In particular, we use the two Intel depth Realsense cameras D435i for both setups. See Figure \ref{fig:hardware} for more details on the robot setup.

\begin{table*}
\centering
\begin{tabularx}{
\textwidth}{@{}lc*{2}{>{\centering\arraybackslash}X}@{}}
\toprule
Scenario name & Generalist policy success rate & Imitation Learning success rate baseline\\
 \midrule
\addlinespace
 Dim lighting scenario & 30.0$\%$ & 0.0$\%$ \\
 Messy kitchen scenario & 30.0$\%$ & 10.0$\%$\\
 Human disturbance scenario & 50.0$\%$ & 0.0$\%$\\

\midrule
\addlinespace

\bottomrule
\end{tabularx}
\vspace{0.05cm}
\caption{Success rate for various disturbance and distractor scenarios.}
\label{tab:robustness_2}
\end{table*}

\begin{figure}[t!]
    \centering
    \includegraphics[width=\linewidth]{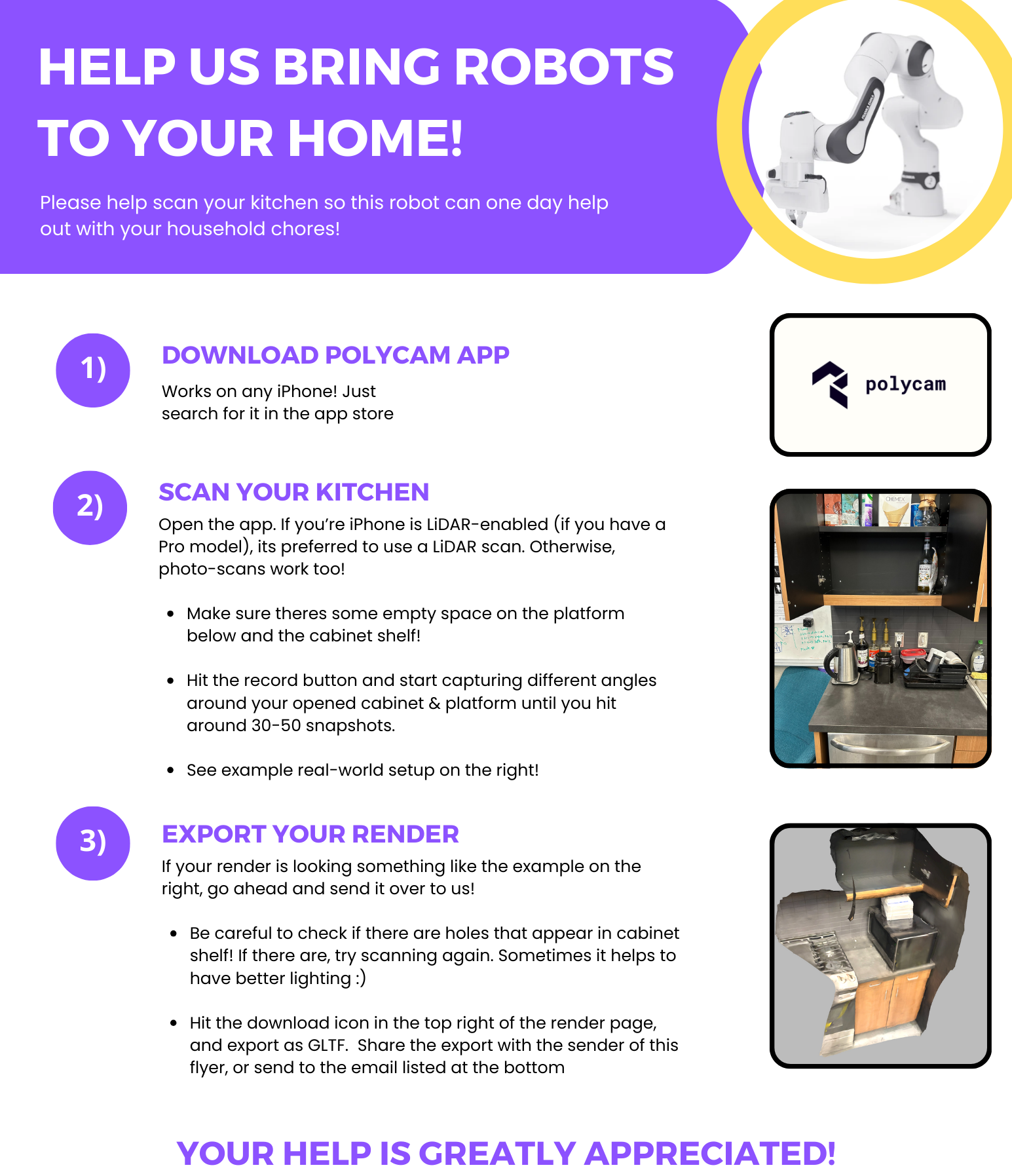}
    \caption{\footnotesize{Poster used for calling crowdsourcing contribution.} }
    \label{fig:crowdsourcing_1}
    \end{figure}
    \begin{figure}
    \includegraphics[width=\linewidth]{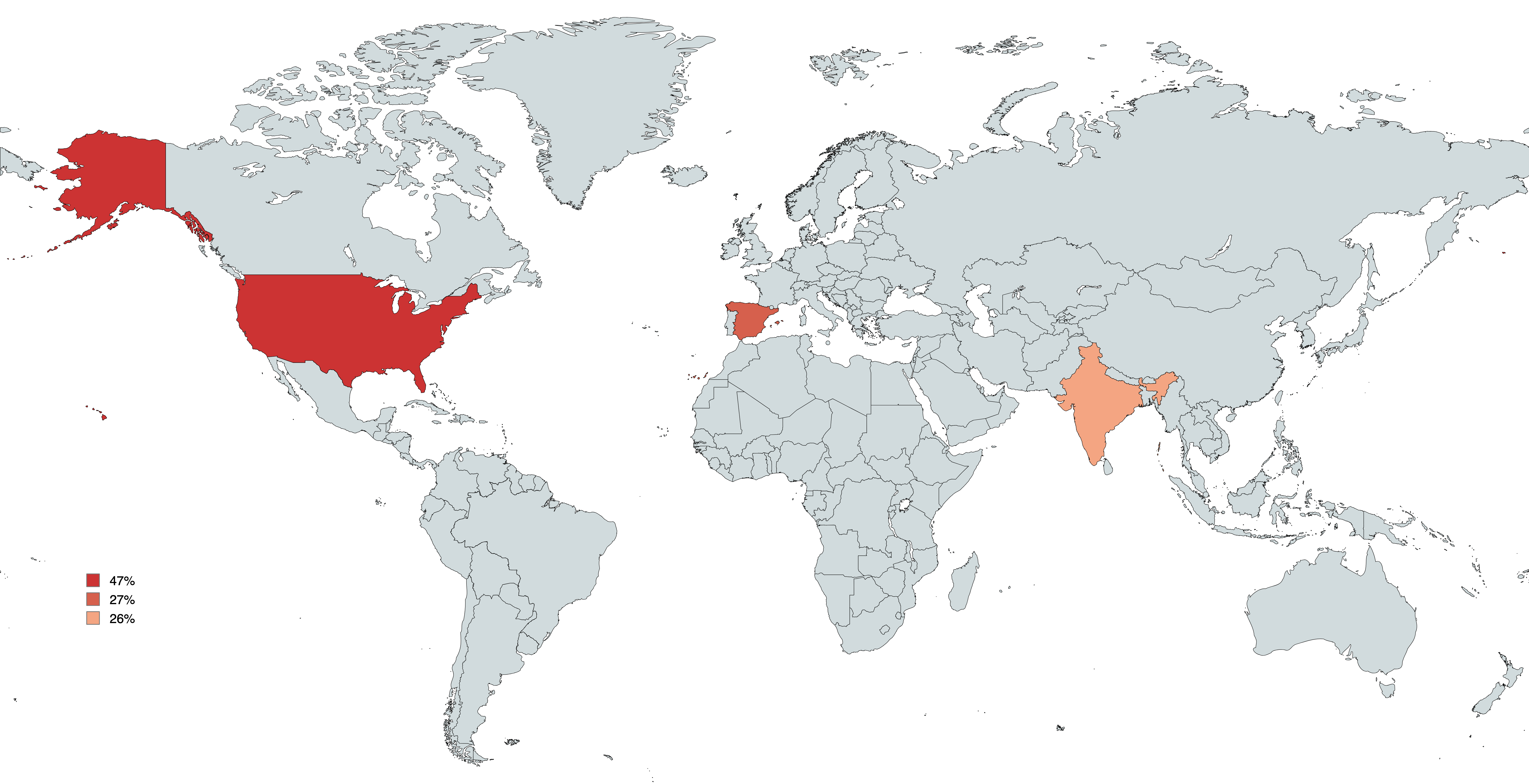}
    \caption{\footnotesize{Geographical distribution of crowdsourcing contributors.} }
    \label{fig:crowdsourcing_2}
\end{figure}

\section{Crowdsourcing}
\label{apdx:crowdsourcing}
We source the kitchen scans from both expert and non-expert users. For placing the object-to-sink task, we collected policies on 29 sink scenes, of which 22 were collected through crowdsourcing. For putting the object to cabinet task, we collect policies on 26 cabinet scenes, of which 18 are collected through crowdsourcing. Figure \ref{fig:crowdsourcing_1} shows the poster we use for crowdsourcing and Figure \ref{fig:crowdsourcing_2} shows the geographical distribution of the crowdsourcing contributors.

\section{Compute Resources}
\label{apdx:compute}
We run all the experiments on an NVIDIA GeForce RTX 3090, an NVIDIA GeForce RTX 3080, and an NVIDIA RTX A6000. The first step of RL fine-tuning is to use the GUI to create a task environment from a crowdsourced kitchen scan and collect a set of 10 demonstrations in simulation using teleoperation, which takes 1 hour per environment on average. We leverage a distributed research computing cluster to run the RL fine-tuning, where we request an NVIDIA Quadro RTX 6000, and it takes an average of 20 hours to converge. Finally, during the teacher-student distillation step, it takes 4 hours on average to collect the simulation trajectories 2 hours to collect the synthetic trajectories, and 5 days to distill into the vision policy.

\end{document}